\documentclass[dvipsnames,format=sigconf,anonymous=false,review=false,nonacm]{acmart}
\AtBeginDocument{%
  }

\setcopyright{acmcopyright}
\copyrightyear{2023}
\acmYear{2023}

\acmDOI{10.1145/nnnnnnn.nnnnnnn} % To be updated after completing copyright process
\acmISBN{978-x-xxxx-xxxx-x/YY/MM} % To be updated after completing copyright process

\acmConference[GECCO '23]{The Genetic and Evolutionary Computation Conference 2023}{July 15--19, 2023}{Lisbon, Portugal}

\usepackage{booktabs} % for prettier tables
\usepackage{threeparttable} % for tables with footnotes

\usepackage{indentfirst}

\usepackage{listings}
\usepackage{xcolor}
\definecolor{codegreen}{rgb}{0,0.6,0}
\definecolor{codegray}{rgb}{0.5,0.5,0.5}
\definecolor{codepurple}{rgb}{0.58,0,0.82}
\definecolor{backcolour}{rgb}{0.95,0.95,0.92}
\lstdefinelanguage{rock}{
    basicstyle=\footnotesize,
    backgroundcolor=\color{backcolour},   
    commentstyle=\color{codegreen},
    keywordstyle=\color{magenta},
    numberstyle=\tiny\color{codegray},
    stringstyle=\color{codepurple},
    basicstyle=\ttfamily\footnotesize,
    breaklines=true,                 
    keepspaces=false,                 
    numbers=left,                    
    showspaces=false,                
    showstringspaces=false,
    morekeywords={minizinc_task, while, True, llm_completion, llm_edit, break, return, def, str, optimisation_language_task, optimisation_model_generator, auto_fixed_model, solver},
    deletekeywords={input},
    morestring=*[d]{"},
    frame=shadowbox,
}

\usepackage{academicons}

\begin{document}

%\title{Towards an Automatic Optimisation Model Generator Assisted with Large Language Models}

%\title[Auto Optimisation Model by GPT-3.5]{Towards an Automatic Optimisation Model Generator Assisted with Generative Pre-trained Transformer}
\title{Towards an Automatic Optimisation Model Generator Assisted with Generative Pre-trained Transformer}

\author{Boris Almonacid}
\email{boris.almonacid@globalchange.science}
\orcid{0000-0002-6367-9802}
\affiliation{%
  \institution{Global Change Science}
  %\streetaddress{P.O. Box 1212}
  \city{Puerto Varas}
  \country{Chile}
}

\renewcommand{\shortauthors}{Boris Almonacid}

\begin{abstract}
This article presents a framework for generating optimisation models using a pre-trained generative transformer. The framework involves specifying the features that the optimisation model should have and using a language model to generate an initial version of the model. The model is then tested and validated, and if it contains build errors, an automatic edition process is triggered. An experiment was performed using MiniZinc as the target language and two GPT-3.5 language models for generation and debugging. The results show that the use of language models for the generation of optimisation models is feasible, with some models satisfying the requested specifications, while others require further refinement. The study provides promising evidence for the use of language models in the modelling of optimisation problems and suggests avenues for future research.
\end{abstract}






% from: https://shantoroy.com/latex/acm-remove-copyright-information-from-first-page/
%\settopmatter{printfolios=true}
\maketitle

\par\null
\par\null
%\par\null

\newcommand\blfootnote[1]{%
  \begingroup
  \renewcommand\thefootnote{}\footnote{#1}%
  \addtocounter{footnote}{-1}%
  \endgroup
}

\definecolor{orcidlogocol}{HTML}{A6CE39}

\section{Introduction}

\blfootnote{
Boris Almonacid ORCID: \url{https://orcid.org/0000-0002-6367-9802} \\
Preprint log:\\
- 14 Apr 2023: Submitted to Genetic and Evolutionary Computation Conference (GECCO-2023) as Late-Breaking Abstract. \\
- 03 May 2023: Reject by GECCO-2023.\\
- 09 May 2023: Submitted to arXiv.
}

Optimisation models play a critical role in many industries, including healthcare, energy, food industry, and transportation. However, creating an accurate and efficient model can be a time-consuming task as you have to learn modelling languages. To address this challenge, this paper proposes an automatic optimisation model generator that is assisted by a pre-trained generative transformer (GPT) \cite{vaswani2017attention}.
Recent advances in natural language processing \cite{chen2021evaluating} have led to the development of powerful large language models, such as GPT-3, that can generate high-quality text.
The use of GPT for software error correction has been demonstrated in prior work \cite{sobania2023analysis}, which provides a basis for its potential application in other domains, including optimisation problem modelling.
Taking advantage of the capabilities of these models and progress, this research aims to automate the optimisation model generation process by initially defining the necessary features of the problem and using the language model to generate an initial version of the model or repair it in case of finding errors.
This approach has the potential to significantly improve the speed and accuracy of optimisation model generation, making it possible to quickly generate models that meet desired specifications. Additionally, it can help reduce the expertise required to build these models, making the process accessible to a broader audience.

%\newpage
\section{The Method}

\begin{figure}[h]
\centering
\includegraphics[width=\linewidth]{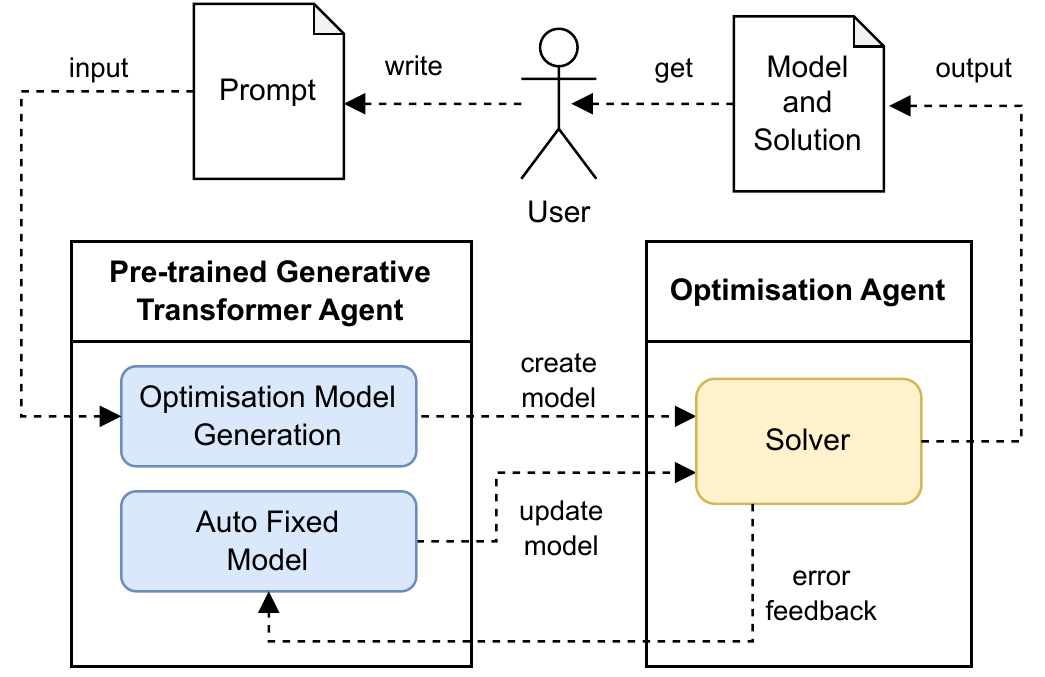}
\caption{Outline of the Automatic Optimisation Model Generator Assisted with a Pre-trained Generative Transformer.}
\label{figure:outline}
\end{figure}

%El usuario escribe un promt indicando las características que debe de tener el modelo de optimsiación. Estas instrucciones son el input del Large Language Model (LLM) Agent  que es el encargado de generar el modelo de optimisación. Una vez que el modelo es creado se envia al agente de optimización ell cual tiene la función de compilar y ejectutar el modelo de optimisación.
%En el caso que que el agente de optimisación compile el modelo y lo resuelva satisfactoriamente, se le entregan como resultados al usuario  el modelo de optimisación y la solución.
%The event in which the optimisation model contains compilation errors trigger an automatic fixed process, whereby the error message is used to provide feedback and resolve problems in the model.
The approach is outlined in Figure \ref{figure:outline}.
%The user writes a prompt indicating the features that the optimisation model must have.
The user specifies the desired features of the optimisation model through a prompt.
These instructions are the input of the GPT Agent, which is responsible for generating the optimisation model. Once the model is created, it is sent to the optimisation agent, which has the function of compiling and executing the optimisation model. In the event that the optimisation agent compiles the model and solves it satisfactorily, the optimisation model and the solution are provided to the user as a result. The event in which the optimisation model contains compilation errors triggers an automatic fixed process, whereby the error message is used to provide feedback and resolve problems in the model. Algorithm \ref{algorithm:llm_main} describes the steps mentioned above.

%Build the indicator considering the features that the optimisation model to be generated must have.

%Execute the instruction in the large language model agent. In this step, the first version of the optimization model is obtained.

%Run the model in the optimisation agent.

\par\null
\par\null
\par\null

\begin{lstlisting}[caption=Automatic Optimisation Model Generator, label=algorithm:llm_main, language=rock]
instruction = "Me: A source code with 10 discrete
               variables." 
model = optimisation_model_generator(instruction)
while True:
  status, output = solver(model)
  if status == True:
    return output
  else:
    input = model
    instruction = "Me: Fix the minizinc code. The Error
                   code is " + output +  "Bot:"
    model = auto_fixed_model(input, instruction)
\end{lstlisting}

\section{Experiment}
%\section{Experiment and Discussion}

% Se ha utilizado el lenguaje de modelado MiniZinc como el lenguaje objetivo para la generación de los modelos de optimisación.
The MiniZinc\footnote{\ \ The version used was MiniZinc 2.7.1, https://www.minizinc.org} modelling language \cite{nethercote2007minizinc} has been used as the target language for the generation of optimisation models.
The models were solved by Gecode 6.3.0.
%Para la generación de las pruebas se utilizan dos modelos de lenguage de GPT-3.5
The experimentation protocol involves the utilisation of two distinct GPT-3.5 language models\footnote{\ \ https://platform.openai.com/docs/models/gpt-3-5} for the generation of the tests.
%para la generación de un modelo inicial de optimisación se utilizará el modelo text-davinci-003 es un modelo de lenguaje de gran tamaño que utiliza una combinación de técnicas de redes neuronales para generar texto coherente y bien estructurado.
For the generation of an initial optimisation model, the \textsc{text-davinci-003} model will be used, it is a large language model that uses a combination of neural network techniques to generate coherent and well-structured text.
%Por otro lado, se utilizará el modelo text-davinci-edit-001 que se enfoca en la edición de texto y corrección de errores.
On the other hand, the \textsc{text-davinci-edit-001} model\footnote{\ \ https://openai.com/blog/gpt-3-edit-insert} will be used that focuses on text editing and error correction.
%text-davinci-edit-001, por otro lado, es un modelo de lenguaje más pequeño que se centra en la edición de texto en lugar de la generación de texto desde cero. Este modelo está diseñado para corregir errores gramaticales y ortográficos en el texto de entrada y puede ser útil para tareas como la corrección automática de textos.
%En resumen, text-davinci-003 se enfoca en la generación de texto de alta calidad, mientras que text-davinci-edit-001 se enfoca en la edición de texto y corrección de errores.
% Los parámetros de   ... estan descritos las funciones del algorithm  
The parameters used in the functions for code generation and automatic code editing are described in the Algorithm \ref{algorithm:llm_functions}.

%\footnote{\ \ https://platform.openai.com/docs/guides/completion/prompt-design}

\begin{lstlisting}[caption=Parameters of GPT functions, label=algorithm:llm_functions, language=rock]
def optimisation_model_generator(instruction: str):
    response = openai.Completion.create(model="text-davinci-003", prompt=instruction, temperature=0, max_tokens=200, top_p=1, frequency_penalty=0.0, presence_penalty=0.0, stop=["Bot:", "Me:"])
    return response.choices[0].text, response

def auto_fixed_model(input: str, instruction: str):
    response = openai.Edit.create(model="text-davinci-edit-001", input=input, instruction=instruction, temperature=0, top_p=1)
    return response.choices[0].text, response
\end{lstlisting}

%La prueba involucra 10 escenarios o instancias, siendo los primeros 5 que implican variables discretas y los últimos 5 que involucran matrices de variables discretas. Para cada escenario, la prueba considera si el dominio de la variable es abierto o definido, si se aplica o no una restricción.
The test involves 10 instances, with the first 5 involving discrete variables and the last 5 involving a matrix composed of discrete variables. For each instance, the test considers whether the domain of the variable is open or defined, and whether or not a constraint applies.
%Estas instancias seran ingresadas en el sistema mediante un prompt que indica esas caracteristicas.
These instances will be entered into the system through a prompt\footnote{\ \ ``Bot: Ask me any questions about the MiniZinc. MiniZinc is a high-level constraint programming language used for modelling and solving combinatorial optimisation problems.
Me: Can I ask you about codes written in MiniZinc as an example? Can you show only the source code?
Bot: Yes. Tell me what kind of language optimisation problems MiniZinc would like me to generate for you.
Me: A source code with 10 discrete variables without domain and without constraints. Put the Bot comments with \% symbol.
Bot:''} indicating these features.
% La validez del modelo de optimisacion es realizado en tiempo de ejecución por la librería de MiniZinc python. 
The MiniZinc Python library\footnote{\ \ https://github.com/MiniZinc/minizinc-python} is used for run-time validation of the optimisation model. 
%Para verificar si el modelo de optimisacion obtenido es correcto a las instrucciones del prompt, se realiza una inspección manual del código fuente del modelo.
In order to ensure whether the obtained optimisation model is correct according to the instructions in the prompt, a manual inspection of the generated source code is performed.

% Los resultados de las pruebas se encuentran descritos en la tabla 1.
The results of the tests are described in Table \ref{table:results}. 
% La columna Variable y Domain indica que pruebas son discretos con o sin dominio,
%, y el paso requerido para completar la prueba. También se indica el número de tokens necesarios para completar cada escenario.
%Los modelos generados para las instancias 1, 2, 3, y 4 han sido generadas validamente y correctamente los modelos de opptimización.
%The optimisation models for instances 1, 2, 3, 4, and 9 have been validly and correctly generated.
%For instance 5 and 10 the constraint is added to the model, but the "alldifferent.mzn" library is not added.
%The model for instances 5 and 10 includes the constraint, but the \textsc{alldifferent.mzn} library is omitted from both models.
% Se han generado modelos válidos para las instancias 6, 7, y 8, sin embargo, los modelos no son correcto con lo solicitado.
%Models have been generated for instances 6, 7 and 8 that are valid. However, the models do not meet the requested specifications.
The results indicate that the use of GPT-3.5 language models for the generation of optimisation models in MiniZinc is feasible. The models generated for instances 1, 2, 3, 4, and 9 were valid and met the requested specifications. However, the models generated for instances 5 and 10 despite including the \textsc{all\_different} constraint, the \textsc{include "alldifferent.mzn";} library was omitted from both models. 
%En una inspección visual, el mensaje de error no indica que falta una librería, sino que falta un predicado o función.
Models have been generated for instances 6, 7 and 8 that are valid. However, the models do not meet the requested specifications.

\begin{table}
\centering
\small
\begin{threeparttable}
\caption{Results of the tests in the automatic generation of MiniZinc models.}
\label{table:results}
\begin{tabular}{clcccccc}
\toprule
ID & Variable    & Domain  & Const.    & Valid  & Correct  & Step & Token \\
\midrule
1  & discrete    & open    & no        & yes & yes & 2    & 508    \\
2  & discrete    & open    & yes       & yes & yes & 2    & 584    \\
3  & discrete    & defined & no        & yes & yes & 1    & 170    \\
4  & discrete    & defined & yes       & yes & yes & 1    & 205    \\
5  & discrete    & defined & all\_diff & no  & no  & 10   & 1712   \\
6  & array disc. & open    & no        & yes & no  & 2    & 293    \\
7  & array disc. & open    & yes       & yes & no  & 2    & 359    \\
8  & array disc. & defined & no        & yes & no  & 2    & 337    \\
9  & array disc. & defined & yes       & yes & yes & 2    & 349    \\
10 & array disc. & defined & all\_diff & no  & no  & 10   & 1787   \\
\bottomrule
\end{tabular}
\begin{tablenotes}
%\item[a] In the case of instances 1-5, ten discrete variables were requested in response to the prompt. For instances 6-10, a single array composed of discrete variables was requested in the prompt.
\item[a] %La columna "valid" indica que el modelo es valido en su ejecución. La columna "correct" señala que el modelo generado corresponde a lo solicitado.
The \textsc{valid} column indicates that the model is valid in its execution. The \textsc{correct} column indicates that the generated model corresponds to what was requested.
\item[b] The data supporting this study's results are available in Figshare at \cite{almonacid_2023}.
\end{tablenotes}
\end{threeparttable}
\end{table}

%Se requiere mensajes de errores más explicitos, centrados en el usuario y en la máquina.

%\section{Conclusions}
\section{Conclusions}

In conclusion, this study provides promising evidence for the use of GPT-3.5 language models in the automatic modelling of optimisation problems. Future research could explore other language models and evaluate their performance compared to the GPT-3.5 models used in this study. Additionally, future studies could explore the use of other optimisation problem modelling languages, in the study of more explicit error messages and user-centered and machine-centered error messages.

\begin{acks}
Boris Almonacid acknowledges the support of PhD (h.c) Sonia Alvarez, Chile. The founders had no role in study design, data collection and analysis, the decision to publish, or the preparation of the manuscript. This study was not externally funded, including by the Chilean Government or any Chilean Universities.
\end{acks}

%\newpage
\bibliographystyle{ACM-Reference-Format}
\bibliography{references}

\end{document}